\pdfoutput=1

\documentclass[11pt]{article}

\usepackage{emnlp2021}

\usepackage{microtype}
\usepackage{amsmath}
\usepackage{booktabs}
\usepackage{array}
\newcolumntype{L}{>{$}l<{$}}
\newcolumntype{C}{>{$}c<{$}}
\newcolumntype{R}{>{$}r<{$}}

\usepackage{times}
\usepackage{latexsym}

\usepackage[T1]{fontenc}

\usepackage[utf8]{inputenc}

\usepackage{microtype}

\usepackage{times}
\usepackage{latexsym}

\usepackage{microtype}

\usepackage{times}
\usepackage{latexsym}

\usepackage[T1]{fontenc}

\usepackage[utf8]{inputenc}

\usepackage{multirow}
\usepackage{array, tabularx, caption, boldline}
\usepackage{graphicx}
\usepackage{cellspace}
\usepackage{algpseudocode}  
\usepackage{amsmath}
\usepackage{multicol}  
\usepackage{multirow} 
\usepackage{flexisym}
\usepackage{graphicx}  
\usepackage{array}

\usepackage{microtype}

\usepackage{tabularx}
\makeatletter
\def\hlinewd#1{%
\noalign{\ifnum0=`}\fi\hrule \@height #1 %
\futurelet\reserved@a\@xhline}
\makeatother

\usepackage{times}
\usepackage{latexsym}

\usepackage{algpseudocode}  
\usepackage{amsmath}
\usepackage{multicol}  
\usepackage{multirow} 
\usepackage{graphicx}  
\usepackage{array}
\usepackage{arydshln}
\usepackage{booktabs}
\usepackage{textcomp}

\usepackage{amssymb}
\usepackage{pifont}

\usepackage{cleveref}
\crefformat{section}{\S#2#1#3} 
\crefformat{subsection}{\S#2#1#3}
\crefformat{subsubsection}{\S#2#1#3}

\usepackage{microtype}

\usepackage{adjustbox}
\usepackage{multicol}  
\usepackage{multirow} 
\usepackage{amsmath}
\usepackage{amsfonts}
\usepackage{makecell}
\usepackage{algpseudocode} 
\usepackage{verbatim}
\usepackage{mathtools}
\DeclareMathOperator*{\argmax}{arg\,max}

\usepackage{booktabs}

\usepackage{makecell}
\usepackage{cleveref}
\usepackage[normalem]{ulem}

\usepackage{times}
\usepackage{latexsym}

\usepackage{footnote}
\makesavenoteenv{tabular}
\makesavenoteenv{table}

\usepackage[hang,flushmargin]{footmisc}

\usepackage[linesnumbered,ruled]{algorithm2e}
\SetAlFnt{\small}
\SetAlCapNameFnt{\normalsize}
\makeatletter
\newcommand{\nosemic}{\renewcommand{\@endalgocfline}{\relax}}
\newcommand{\dosemic}{\renewcommand{\@endalgocfline}{\algocf@endline}}
\let\oldnl\nl
\newcommand{\nonl}{\renewcommand{\nl}{\let\nl\oldnl}}
\makeatother

\usepackage{enumitem}
\usepackage{tablefootnote}

\usepackage{tabularx}
\makeatletter
\def\hlinewd#1{%
\noalign{\ifnum0=`}\fi\hrule \@height #1 %
\futurelet\reserved@a\@xhline}
\makeatother

\title{Plan-then-Generate: Controlled Data-to-Text Generation via Planning}

\author{Yixuan Su$^{\spadesuit,}$\thanks{~~Work done while the author was an intern at Apple.}~\quad David Vandyke$^{\heartsuit}$\quad Sihui Wang$^{\heartsuit}$\quad Yimai Fang$^{\heartsuit}$\quad Nigel Collier$^{\spadesuit}$\\

$^{\spadesuit}$Language Technology Lab, University of Cambridge \\
$^\heartsuit$Apple\\

{\tt \{ys484,nhc30\}@cam.ac.uk}\\ 
{\tt \{dvandyke,sihui\_wang,yimai\_fang\}@apple.com}
}

\begin{document}
\maketitle
\begin{abstract}
Recent developments in neural networks have led to the advance in data-to-text generation. However, the lack of ability of neural models to control the structure of generated output can be limiting in certain real-world applications. In this study, we propose a novel Plan-then-Generate (PlanGen) framework to improve the controllability of neural data-to-text models. Extensive experiments and analyses are conducted on two benchmark datasets, ToTTo and WebNLG. The results show that our model is able to control both the intra-sentence and inter-sentence structure of the generated output. Furthermore, empirical comparisons against previous state-of-the-art methods show that our model improves the generation quality as well as the output diversity as judged by human and automatic  evaluations.
\end{abstract}

\section{Introduction}
Generating natural language from structured data \cite{DBLP:journals/jair/GattK18}, i.e. data-to-text generation, is a research problem that is crucial to many downstream NLP applications. Some examples are dialogue systems \cite{DBLP:conf/naacl/WenGMRSVY16}, restaurant assistant \cite{DBLP:conf/sigdial/NovikovaDR17}, and open domain question answering \cite{DBLP:journals/corr/abs-2010-10439}.

To address this task, many researchers have designed sophisticated neural models based on various methods, such as soft-template \cite{DBLP:conf/emnlp/WisemanSR18}, copy mechanism \cite{DBLP:conf/inlg/GehrmannDER18}, and pre-trained language models 
\cite{DBLP:journals/corr/abs-2005-10433,DBLP:journals/corr/abs-2007-08426}. While achieving impressive results, most existing studies only focused on producing results that are close to the references. On the other hand, the controllability of such models is still under-explored, i.e. what to generate and in what order (the output structure) in their outputs cannot be explicitly controlled by the users.\\\indent We argue that the model's ability to control the structure of its output is highly
\begin{table}[t] 
	\centering    
	\setlength{\abovecaptionskip}{3pt}
\includegraphics[width=0.43\textwidth]{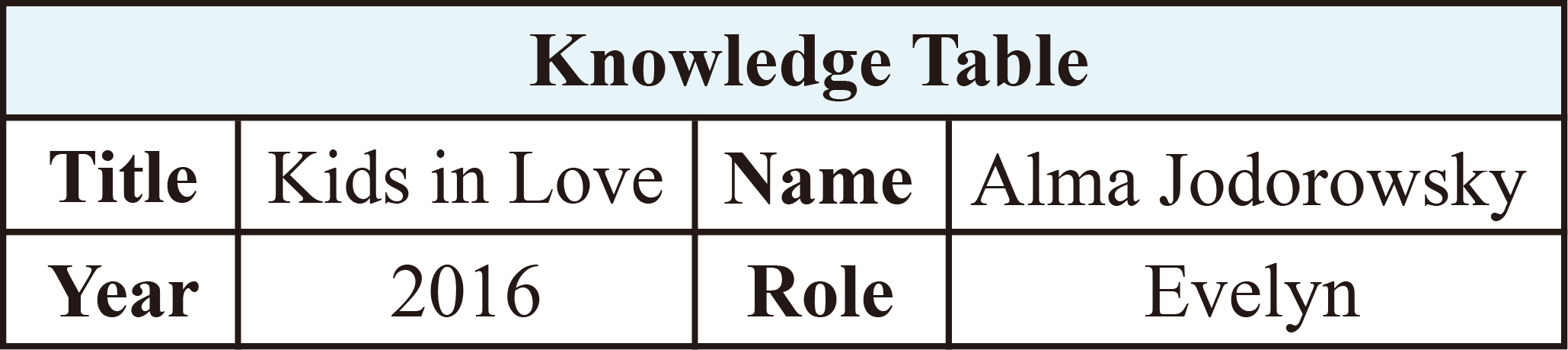}
	\caption{An Example of Knowledge Table}
    \label{fig:table_example}
\end{table}
desirable for at least two reasons. (1) Arranging the structure of the output in a certain form enables it to have greater naturalness, as the structure of the sentence often reflects the salience of the entities it contains \cite{DBLP:journals/coling/PoesioSEH04}. Suppose we have a digital assistant which replies to user queries based on knowledge tables like Table \ref{fig:table_example}. Then, for a user query ``\textit{Who played Evelyn in Kids in Love?}'', a natural response is ``\textit{Evelyn in Kids in Love was played by Alma Jodorowsky.}''. In contrast, to a different query ``\textit{What role did Alma Jodorowsky play in Kids in Love?}'', a natural response would be ``\textit{Alma Jodorowsky played Evelyn in Kids in Love.}''. 
While both answers are semantically equivalent, producing the answer with the most appropriate structure allows the system to sound less robotic and be easily understood. (2) It allows the model to generate outputs with diverse structures by simply changing the input planning information (i.e. a content plan), which could potentially benefit other applications such as paraphrasing and data augmentation. \\\indent
To control the output structure, we need an intermediate ``planning''
\begin{figure*}[t] 
	\centering    
	\setlength{\abovecaptionskip}{3pt}
\includegraphics[width=1.0\textwidth]{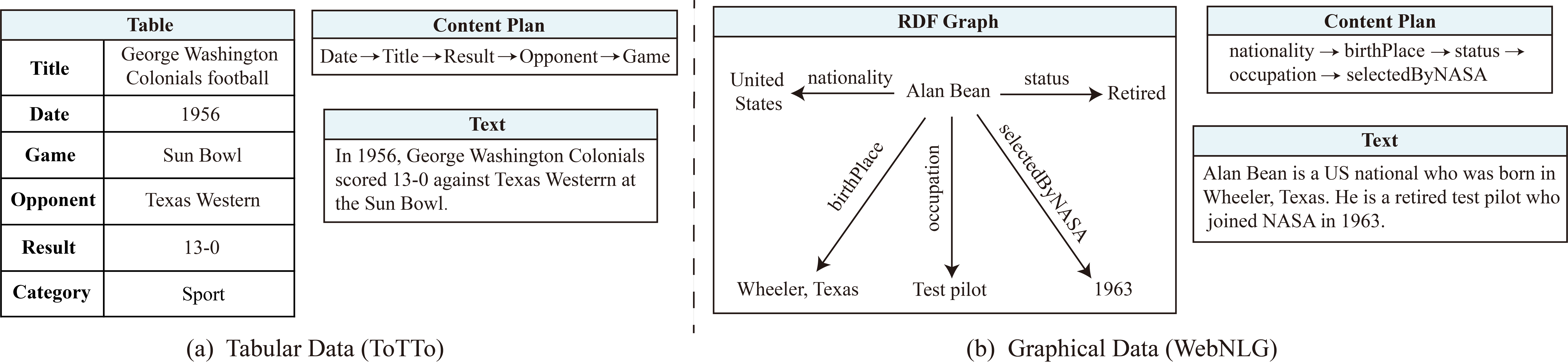}
	\caption{Plot illustrating the relationship between the structured data, the content plan, and the reference text for examples from (a) ToTTo dataset (tabular data) and (b) WebNLG dataset (graphical data with RDF structure).}
    \label{fig:plan_example}
\end{figure*}
signal (i.e. a content plan) which informs the model what to generate and in what order. To this end, we propose a Plan-then-Generate (PlanGen) framework which consists of two components: a content planner and a sequence generator. Given the input data, the content planner first predicts the most plausible content plan that the output should follow. Then, the sequence generator takes the data and the content plan as input to generate the result. To further ensure the controllability of our model, we propose a structure-aware reinforcement learning  (RL) objective that encourages the generated output to adhere to the given content plan. 
In this work, we formulate the intermediate content plan as an ordered list of tokens for its simplicity and wide applicability to data with different structures. For tabular data, each token in the content plan is a slot key from the table. As for graphical data with RDF structure, each token represents the predicate from an RDF triple. In Figure \ref{fig:plan_example}, we provide examples for both cases. 

To fully evaluate our approach, we test the proposed model on two benchmarks with different data structures: (i) ToTTo dataset \cite{DBLP:conf/emnlp/ParikhWGFDYD20} with tabular data, and (ii) WebNLG dataset \cite{DBLP:conf/inlg/ColinGMNP16,DBLP:conf/inlg/GardentSNP17} with graphical data. Compared with previous state-of-the-art approaches, our model achieves better performance in terms of generation quality as judged by both human and automatic evaluations. In particular, the results also show that the outputs of our model are highly controllable and contain diverse structures.



In summary, our contributions are: (1) A novel Plan-then-Generate (PlanGen) framework that consists of a content planner and a sequence generator for data-to-text generation. (2) Extensive automatic and human evaluations reporting state-of-the-art results on two benchmark datasets. (3) In-depth analysis revealing the merits of the proposed approach in terms of controllability and diversity. 

\section{Related Work}
Data-to-text generation is a long-standing problem \cite{DBLP:journals/nle/ReiterD97} that aims at producing natural language descriptions of structured data. Traditional systems are primarily built on template-based algorithms \cite{oh-rudnicky-2000-stochastic,stent-etal-2004-trainable,DBLP:conf/acl/KondadadiHS13}. With recent advances in deep learning, researchers have shifted their attention to neural generation models that can be summarized into two categories. 
\begin{figure*}[t] 
	\centering    
	\setlength{\abovecaptionskip}{3pt}
\includegraphics[width=0.97\textwidth]{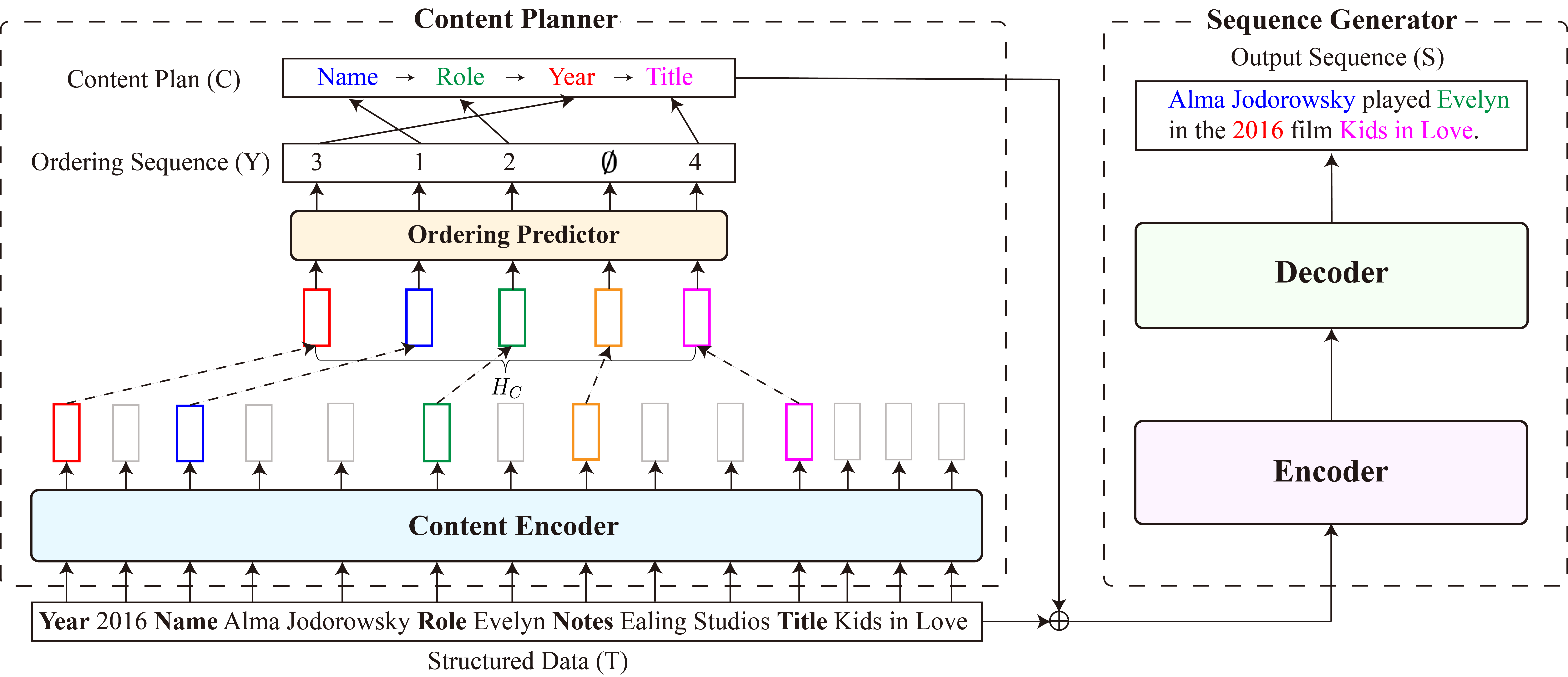}
    \caption{\textbf{PlanGen Framework}: Given the structured data ($T$), a content plan ($C$) is first predicted by the content planner (left). The sequence generator (right) then takes the structured data and the predicted content plan as input to generate the output ($S$). Note that, the content plan can also be specified by the user for a controlled generation.}
    \label{fig:overview}
\end{figure*}

\paragraph{End-to-End Models.} Many existing studies are dedicated to building end-to-end neural models with different strategies like soft-templates \cite{DBLP:conf/emnlp/WisemanSR18,DBLP:conf/iclr/YeS0W020}, attention awareness \cite{DBLP:conf/aaai/LiuWSCS18,DBLP:conf/inlg/ColinG19}, and retrieved prototypes \cite{DBLP:conf/ijcai/LiLDZS20,su2021fewshot}. \citet{DBLP:conf/inlg/GehrmannDER18}, \citet{DBLP:conf/aaai/Puduppully0L19,DBLP:conf/acl/PuduppullyDL19}, and \citet{DBLP:conf/acl/ChenECLW20} adopted copy mechanism for content selection to improve the information coverage of the outputs. With recent advance in pre-trained language models (PLMs) \cite{DBLP:conf/naacl/DevlinCLT19,DBLP:journals/corr/abs-1907-11692,DBLP:journals/jmlr/RaffelSRLNMZLL20,DBLP:conf/acl/LewisLGGMLSZ20}, several researchers \cite{DBLP:conf/emnlp/ChenSYW20,DBLP:conf/acl/ChenECLW20,DBLP:journals/corr/abs-2005-10433,DBLP:journals/corr/abs-2007-08426} have studied the ways to adapt PLMs into the data-to-text generation task. 
\paragraph{Pipeline Models.}  Another line of research investigates ways to tackle the generation problem in a pipeline framework. \citet{DBLP:conf/acl/MaYLLZS19} proposed to first use a classifier to select the key contents. The planning and surface realisation of the selected contents are then addressed by a subsequent Seq2seq model. More related to our work, some researchers studied how neural models can benefit from traditional NLG steps \cite{DBLP:conf/acl/Kukich83,DBLP:books/daglib/0073477}, that is, (i) content planning and (ii) surface realisation. To simultaneously select the key contents and arrange their orderings (i.e. content planning), different strategies are proposed such as the most probable traversal of graph trees \cite{DBLP:journals/corr/abs-1904-03396}, the ordering of graph nodes \cite{DBLP:conf/acl/ZhaoWC20}, and the multi-step pipeline that includes discourse ordering, lexicalization, and regular expression generation \cite{DBLP:conf/emnlp/FerreiraLMK19}. While achieving satisfactory results, these approaches can only be applied to data with graphical structure. 
Compared with previous studies, we show that our content planning approach is more accurate and 
less dependent on the data structure. In addition, by providing the desired content plan, our model can control the output structure on both the intra-sentence and inter-sentence levels (\cref{sec:case_study}).

\section{Preliminaries}
\paragraph{Dataset.} In this study, our training dataset is defined as $\mathcal{D}=\{(T, C, S)_i\}_{i=1}^{|D|}$. (1) $T$ is the linearized structured data and it is defined as $T=\{t_1, ..., t_{|T|}\}$. For data with tabular structure, each item $t_{i}=\{k_i, v_i\}$ is a pair of slot key $k_i$ and slot value $v_i$ (e.g., (\textit{Date}, \textit{1956}) in Figure \ref{fig:plan_example}(a)). As for graphical data with RDF structure, each item $t_{i}=\{s_i, p_i, o_i\}$ represents a RDF triple, where $s_i$, $p_i$, and $o_i$ are subject, predicate, and object, respectively. For instance, in Figure \ref{fig:plan_example}(b), (``Alan Bean'', ``status'', ``Retired'') is a RDF triple. (2) The reference content plan $C$ is defined as $C=\{c_1, .., c_{|C|}\}$, where each token $c_i$ either denotes a slot key (for tabular data) or a predicate (for graphical data). The content plan is thus a selection of the content from the structured data that should appear in the output, in a particular order. (3) The $S=\{s_1, .., s_{|S|}\}$ denotes the reference text. 



\paragraph{Content Plan Construction.}
\label{sec:data_collection}
Note that the original ToTTo and WebNLG datasets only consist of pairs of structured data and reference text. Thus, we use a heuristic delexicalizer $\mathcal{F}$ to construct the reference content plan. For a tabular data $T$, given the reference text $S$, the content plan $C=\mathcal{F}(T, S)$ is built by replacing the parts of the reference text that comes from the table slot values with the corresponding slot keys. For instance, suppose we have a text ``\textit{Alma Jodorowsky played Evelyn in Kids in Love.}'' and Table \ref{fig:table_example}, then the resulting content plan is \{``Name''$\rightarrow$``Role''$\rightarrow$``Title''\}. For graphical data with RDF structure, we apply a similar procedure to build the reference content plan by replacing the parts of the reference text that comes from the objects of the RDF triples with the corresponding predicates. In Figure \ref{fig:plan_example}, we show examples of reference content plan for both cases. 


\section{Methodology}
Figure \ref{fig:overview} depicts the proposed Plan-then-Generate (P2G) framework. Given the input data, the content planner (\cref{sec:content_planner}) first predicts the most probable content plan. The sequence generator (\cref{sec:sequence_generator}) then takes the structured data and the predicted content plan to generate the output. In the following, we elaborate the details of the proposed framework. 

\subsection{Content Planner}
\label{sec:content_planner}
Our content planner consists of two components. The first part is a content encoder which takes the data $T$ as input and produces its representation $H_T\in\mathbb{R}^{|T|\times n}$, where $n$ is the output size. We construct our content encoder with a pre-trained BERT-base model \cite{DBLP:conf/naacl/DevlinCLT19}.


After getting the data representation, we select the hidden states from $H_T$ that corresponds to the tokens\footnote{For tabular data, the selected tokens correspond to all slot keys from the table. Similarly, for graphical data, the selected tokens correspond to the predicates of all input RDF triples.} that might appear in the content plan. Here, we denote the selected hidden states $H_C\in\mathbb{R}^{|C|\times n}$ as $H_C=\{h^c_1, ..., h^c_{|C|}\}$, where $|C|$ is the number of selected tokens from the input data. Next, $H_C$ is fed into the ordering predictor which predicts the orderings of the selected tokens in the predicted content plan. Inspired by \citet{DBLP:conf/eacl/SuCWVBLC21}, we model the ordering predictor as a linear-chain conditional random field (CRF) \cite{DBLP:conf/icml/LaffertyMP01} for its ability to compute the global optimal ordering sequence. When predicting the ordering, the ordering predictor is allowed to emit an empty label $ \emptyset$ which indicates the omission of the corresponding token in the content plan. 

During training, the likelihood of the ordering sequence $Y$ defined by the content plan is 
\begin{align}
\begin{split}
\label{score_function}
    &P_{\textup{CRF}}( Y|H_C)= \frac{e^{f(Y, H_C)}}{\sum_{Y^{\prime}}e^{f(Y^{\prime}, H_C)}}\\
    &=\frac{1}{Z}\exp(\sum_{i=1}^{|C|}\Phi_{y_i}(h^c_i) + \sum_{i=2}^{|C|}M_{y_{i-1}, y_{i}}).\\
\end{split}
\end{align}
Here, $\Phi_{y_i}(h^c_i)$ is the label score of $y_i$ at step $i$, where label $y_i$ indicates the position of the token in the final content plan. Taking Figure \ref{fig:overview} as an example, the position of the ``\textcolor{blue}{Name}'' key is $1$, meaning that ``\textcolor{blue}{Name}'' should appear in the front of the content plan. By predicting the positions instead of the actual slot keys, at test time, our model can handle tables with out-of-vocabulary slot keys that did not appear in the training set. In practice, $\Phi$ is parameterized by a feed-forward layer. The $M_{y_{i-1}, y_{i}}$ denotes the transition score from position $y_{i-1}$ to position $y_i$, and $M$ is a learnable transition matrix. 

During inference, the ordering sequence is predicted as $\Tilde{Y}$ as $\Tilde{Y} = \argmax_{Y^{\prime}}P_{\textup{CRF}}(Y^{\prime}|H_C)$.
As shown in the example of Figure \ref{fig:overview},  given all the slot keys \{``Year'', ``Name'', ``Role'', ``Notes'', ``Title''\} from the table, the predicted ordering sequence is \{3, 1, 2, $\emptyset$, 4\}. The content plan \{``Name'' $\rightarrow$ ``Role'' $\rightarrow$ ``Year'' $\rightarrow$ ``Title''\} can then be predicted by omitting the ``Notes'' key and re-arranging other keys following the predicted ordering sequence.

\subsection{Sequence Generator}
\label{sec:sequence_generator}
Our sequence generator is built on a BART-base model \cite{DBLP:conf/acl/LewisLGGMLSZ20} which consists of a transformer 
based encoder-decoder architecture. 

Given the structured data $T$, the reference content plan $C$, and the reference text $S$, the learning objective of the sequence generator is defined as 
\begin{equation}
    \label{eq:mle}
    \mathcal{L}_{\textup{LM}} = -\sum_{i=1}^{|S|}\log P_{G}(S_i|S_{<i}; E([T:C])),
\end{equation}
where $E$, $G$ are the encoder and decoder, and $[\cdot:\cdot]$ denotes the concatenation operation.

\begin{table*}[t]
    \small
	\centering  
	\renewcommand{\arraystretch}{1.2}
	\scalebox{0.96}{
	\begin{tabular}{cccccccccc}
		\hlinewd{0.75pt}
		{\multirow{2}{*}{\textbf{Model}}}&\multicolumn{3}{c}{\textbf{Overall}}&\multicolumn{3}{c}{\textbf{Overlap}}&\multicolumn{3}{c}{\textbf{Non-Overlap}}\\
		\cmidrule(lr){2-4}
		\cmidrule(lr){5-7}
		\cmidrule(lr){8-10}
		&BLEU&PARENT&BLEURT&BLEU&PARENT&BLEURT&BLEU&PARENT&BLEURT\\
		\hline
		NCP&19.2&29.2&-0.576&24.5&32.5&-0.491&13.9&25.8&-0.662\\
		Pointer-Generator&41.6&51.6&0.076&50.6&58.0&0.244&32.2&45.2&-0.092\\
		BERT-to-BERT&44.0&52.6&0.121&52.7&58.4&0.259&35.1&46.8&-0.017\\
		T5-3B&\textbf{49.5}&58.4&0.230&\textbf{57.5}&62.6&0.351&41.4&54.2&0.108\\
		\hline
		Ours&49.2&\textbf{58.7}&\textbf{0.249}&56.9&\textbf{62.8}&\textbf{0.371}&\textbf{41.5}&\textbf{54.6}&\textbf{0.126}\\
		\hlinewd{0.75pt}
	\end{tabular}}
    \caption{ToTTo test set results: All reported results, including ours, can be found in the official Leaderboard.\footnotemark}
	\label{tb:totto_test}
\end{table*}
\footnotetext{\url{https://github.com/google-research-datasets/ToTTo}}

\subsection{Structure-Aware RL Training}
We note that the structure of the generated sequence can only be accurately measured on the sequence-level, which is not directly optimized by the token-level objective (Eq. \eqref{eq:mle}). Therefore, to encourage the generator to follow the sequence-level structure defined by the content plan, we incorporate reinforcement learning into our training process.

Formally, in training, given the structured data $T$ and the reference content plan $C$, the generator first samples an output sequence $S^{\prime}=(S^{\prime}_1, ..., S^{\prime}_{|S^{\prime}|})$, where $S^{\prime}_t$ is the token sampled at time step $t$. The generator parameters $\theta$ are then updated using the REINFORCE algorithm \cite{DBLP:journals/ml/Williams92} as
\begin{align}
        &\mathcal{L}_{\textup{RL}}=-\mathbb{E}_{S^{\prime}\sim P_{\theta}(T, C)}[R(S, S^{\prime}, T, C)] = \label{eq:rl}\\
        &-R(S, S^{\prime}, T, C)\sum_{i=1}^{|S^{\prime}|}\log P_G(S^{\prime}_i|S^{\prime}_{<i};E([T:C])). \nonumber
\end{align}
The reward function $R(S, S^{\prime}, T, C)$ measures the structure of the sampled sequence $S^{\prime}$ against the input content plan $C$, and its surface form against the reference text $S$ as
\begin{equation}
    R(S, S^{\prime}, T, C) = B(S, S^{\prime}) + B(C, C^{\prime}), 
    \label{eq:reward}
\end{equation}
where $B(\cdot, \cdot)$ is the BLEU score \cite{DBLP:conf/acl/PapineniRWZ02}.  $C^{\prime}=\mathcal{F}(T, S^{\prime})$, and $\mathcal{F}$ is described in \cref{sec:data_collection}. By optimizing Eq. \eqref{eq:rl}, the structure of the output is encouraged to follow the content plan.

\subsection{Learning}
The learning objective of the content planner is $\mathcal{L}_{\textup{CRF}}=-\log P_{\textup{CRF}}$ and $P_{\textup{CRF}}$ is defined in Eq. \eqref{score_function}. For the sequence generator, at the first 10k steps, we train it with $\mathcal{L}_{\textup{LM}}$ as described in Eq. \eqref{eq:mle}. Then, we incorporate the structure-aware RL objective (Eq. \eqref{eq:rl}) and further train the sequence generator with $\mathcal{L}_{\textup{LM}} + \mathcal{L}_{\textup{RL}}$ for 5k more steps.

\section{Experiment Setup}
\subsection{Datasets and Evaluation Metrics}
\label{sub:datasets}
\paragraph{ToTTo Dataset}  \cite{DBLP:conf/emnlp/ParikhWGFDYD20} consists of Wikipedia tables paired with human-written descriptions. Each input is a full table with highlighted cells and the model is required to generate the text that describes the highlighted cells.
Similar to previous studies \cite{DBLP:conf/emnlp/ParikhWGFDYD20,DBLP:journals/corr/abs-2005-10433}, we only use the highlighted cells as the model input. We report the automatic result of BLEU-4, PARENT\footnote{PARENT is a word-overlap based metric that reflects the factual accuracy of the generated text in relation to both the input table and the reference sentence.} \cite{DBLP:conf/acl/DhingraFPCDC19}, and a learnt metric BLEURT \cite{DBLP:conf/acl/SellamDP20}. Note that ToTTo features a hidden test set with two splits: Overlap and Non-Overlap. The Non-Overlap set contains out-of-domain examples. To get the test set result, a submission must be made to the leaderboard.

\paragraph{WebNLG Dataset} is used in the WebNLG challenge \cite{DBLP:conf/inlg/GardentSNP17}. For each data instance, the input is a set of RDF triples from DBPedia and the output is their textual description. The test set of WebNLG features a Seen and Unseen subset. The Unseen subset contains out-of-domain instances. Following previous studies, we report the the automatic result of BLEU and METEOR \cite{DBLP:conf/acl/BanerjeeL05}. 

\subsection{Implementation Details}
Our implementation is based on the Huggingface Library \cite{DBLP:journals/corr/abs-1910-03771}. 
We optimize the model using Adam \cite{DBLP:journals/corr/KingmaB14} with a learning rate of $2$e$-5$ and a batch size of $64$.


\section{Results}
In this section, we report the experimental results.
\subsection{ToTTo Results}
We compare our model with the latest models on ToTTo dataset, including NCP \cite{DBLP:conf/aaai/Puduppully0L19}, Pointer-Generator \cite{DBLP:conf/acl/SeeLM17}, BERT-to-BERT \cite{DBLP:journals/tacl/RotheNS20} and T5-3B \cite{DBLP:journals/corr/abs-2005-10433}. Similar to our model, the later two are also based on pre-trained language models.  

Table \ref{tb:totto_test} lists the results on ToTTo test set. For most of the metrics, our model with 140M parameters outperforms the current state-of-the-art T5-3B model which has over 2.8B parameters. The results on the PARENT metric suggest that our model can generate more factually accurate text. Moreover, in the Non-Overlap subset, our model achieves the best result on all metrics, showing its robustness to out-of-domain examples.

\subsection{WebNLG Results}
We compare our approach with two types of models on WebNLG dataset. The first type of models does not use pre-trained language models (PLMs), including GTR-LSTM \cite{DBLP:conf/acl/WangZQT18}, Transformer \cite{DBLP:conf/emnlp/FerreiraLMK19}, Step-by-Step \cite{DBLP:journals/corr/abs-1904-03396}, and PLANENC \cite{DBLP:conf/acl/ZhaoWC20}. Similar to ours, the latter three are pipeline models that utilize different methods to decide the output planning before generating the result. The second line of research utilizes PLMs, including 
Switch-GPT \cite{DBLP:conf/acl/ChenECLW20}, T5 \cite{DBLP:journals/corr/abs-2005-10433}, and T5+Prefix \cite{DBLP:journals/corr/abs-2007-08426}. The Switch-GPT model applies a copy mechanism to copy content from the source to the output. We also include the top systems of the WebNLG challenge, including ADAPT, TILB-SMT, and MELBOURNE. 

\paragraph{Evaluation on Text Generation.} Table \ref{tb:webnlg} lists the results of different methods in terms of text generation. We see that our approach outperforms all prior works. 
Compared with previous models that utilize PLMs, our performance improvements suggest that the incorporation of an explicit content plan can provide effective guiding signal for the model to achieve better generation results.


\begin{table}[t]
    \small
	\centering  
	\renewcommand{\arraystretch}{1.2}
	\scalebox{0.86}{
	\begin{tabular}{ccccccc}
		\hlinewd{0.75pt}
		{\multirow{2}{*}{\textbf{Model}}}&\multicolumn{2}{c}{\textbf{Seen}}&\multicolumn{2}{c}{\textbf{Unseen}}&\multicolumn{2}{c}{\textbf{Overall}}\\
		\cmidrule(lr){2-3}
		\cmidrule(lr){4-5}
		\cmidrule(lr){6-7}
		&B.&M.&B.&M.&B.&M.\\
		\hlinewd{0.75pt}
		ADAPT$^\dagger$&60.59&0.44&10.53&0.19&31.06&0.31\\
		TILB-SMT$^\dagger$&54.29&0.42&29.88&0.33&44.28&0.38\\
		MELBOURNE$^\dagger$&54.52&0.41&33.27&0.33&45.13&0.37\\
		\hline
		GTR-LSTM$^\dagger$ &54.00&0.37&29.20&0.28&37.10&0.31\\
		Transformer$^\dagger$ &56.28&0.42&23.04&0.21&47.24&0.39\\
		Step-by-Step$^\dagger$&53.30&0.44&38.23&0.34&47.24&0.39\\
		PLANENC$^\dagger$&64.42&0.45&38.23&0.37&52.78&0.41\\
		\hline
		\textit{Based on PLMs}&&&&&&\\
		\hline
		Switch-GPT&60.98&0.43&40.67&0.34&52.17&0.40\\
		T5$^\ddagger$&63.90&0.46&52.80&0.41&57.10&0.44\\
		T5+Prefix$^\ddagger$&64.71&0.45&53.67&0.42&59.70&0.44\\
		\hline
		Ours&\textbf{65.42}&\textbf{0.48}&\textbf{54.52}&\textbf{0.44}&\textbf{60.51}&\textbf{0.46}\\
		\hlinewd{0.75pt}
	\end{tabular}}
    \caption{Text generation results on WebNLG datasets, where B. and M. represent BLEU and METEOR metrics. $^\dagger$ and $^\ddagger$ results are cited from \citet{DBLP:conf/acl/ZhaoWC20} and \citet{DBLP:journals/corr/abs-2007-08426}, respectively.}
	\label{tb:webnlg}
\end{table}

\begin{table}[t]
    \small
	\centering  
	\renewcommand{\arraystretch}{1.2}
	\scalebox{0.87}{
	\begin{tabular}{ccccccc}
		\hlinewd{0.75pt}
		{\multirow{2}{*}{\textbf{Model}}}&\multicolumn{2}{c}{\textbf{Seen}}&\multicolumn{2}{c}{\textbf{Unseen}}&\multicolumn{2}{c}{\textbf{Overall}}\\
		\cmidrule(lr){2-3}
		\cmidrule(lr){4-5}
		\cmidrule(lr){6-7}
		&Acc.&B-2&Acc.&B-2&Acc.&B-2\\
		\hlinewd{0.75pt}
		Transformer$^\dagger$ &0.56&74.30&0.09&20.90&0.34&49.30\\
		GRU$^\dagger$&0.56&75.80&0.10&25.40&0.35&52.20\\
		Step-by-Step$^\dagger$&0.49&73.20&0.44&68.00&0.47&70.80\\
		PLANENC$^\dagger$&0.63&80.80&0.61&79.30&0.62&80.10\\
		\hline
		Ours&\textbf{0.74}&\textbf{86.01}&\textbf{0.70}&\textbf{83.79}&\textbf{0.72}&\textbf{84.97}\\
		w/o CRF&0.67&82.92&0.63&80.65&0.65&81.73\\
		w/o PLMs&0.70&84.05&0.65&81.98&0.68&83.02\\
		\hlinewd{0.75pt}
	\end{tabular}}
    \caption{Evaluation results on content planning. $^\dagger$ results are copied from \citet{DBLP:conf/acl/ZhaoWC20}.}
	\label{tb:planner}
\end{table}

\paragraph{Evaluation on Content Planning.} Next, we compare our content planner with other pipeline models in terms of content planning performance. Following \citet{DBLP:conf/acl/ZhaoWC20}, we report the results on planning accuracy (P-A) and planning BLEU-2 score (B-2) against the human-generated plans\footnote{The human-generated plans are provided in the enriched WebNLG dataset \cite{DBLP:conf/inlg/FerreiraMKW18}.}. In addition, we examine two ablated variants of our content planner by either removing the CRF layer (w/o CRF) or using randomly initialized parameters instead of the pre-trained BERT (w/o PLMs). Table \ref{tb:planner} lists the results. We see that our content planner outperforms all the baselines on both measures. Moreover, the results show that both the CRF layer and the pre-trained parameters positively contribute to the overall performance which further justifies our design of the content planner.

\subsection{Human Evaluation}
We also conduct a human evaluation to assess our model, using graders proficient in English from an internal grading platform. We randomly selected 200 samples from the ToTTo validation set. For each sample, we first use our sequence generator to produce the result with the content plan (CP) predicted by the content planner. Next, we randomly shuffle the predicted content plan and generate five different results (Shuffled CP). For comparison, we also include results of BERT-to-BERT and T5-3B using greedy decoding. All generated results, plus the reference sentence, are evaluated by three graders on a 3-point Likert scale (0, 1, or 2) for each of the following features\footnote{More evaluation details are provided in the Appendix \ref{sec:human_evaluation_detail}.}:
\begin{itemize}[noitemsep]
    \item \textbf{Faithfulness}: Whether the sentence is factually consistent with the input data.
    \item \textbf{Fluency}: Whether the sentence is fluent and easy to understand. 
    \item \textbf{Accuracy}: How accurately the sentence follows the input content plans\footnote{As BERT-to-BERT and T5-3B do not take the content plan as input, thus we do not report their accuracy score.}.
\end{itemize}

\begin{table}[tb]
    \small
	\centering  
	\renewcommand{\arraystretch}{1.2}
	\scalebox{0.88}{
	\begin{tabular}{cccc}
		\hlinewd{0.75pt}
        &\textbf{Faithfulness}&\textbf{Fluency}&\textbf{Accuracy}\\
        \hline
        Agreement&0.663&0.617&0.518\\
        \hline
        Reference&1.819&1.762&1.753\\
        \hline
        BERT-to-BERT&1.589&1.593&-\\
        T5-3B&1.701&1.696&-\\
        \hline
        \textbf{Ours}(CP)&1.794&1.753&1.742\\
        \textbf{Ours}(Shuffled CP)&1.778&1.746&1.552\\
		\hlinewd{0.75pt}
	\end{tabular}}
    \caption{Human Evaluation Results}
	\label{tb:human_evaluation}
\end{table}


Table \ref{tb:human_evaluation} lists the results, with the first row showing strong inter-annotator agreements as measured by Fleiss$\textprime$ kappa coefficient \cite{fleiss1971mns}.
Comparing with BERT-to-BERT and T5-3B, our model achieves best results on both measures. Furthermore, on the faithfulness and fluency metrics, our model with both CP and Shuffled CP performs comparably with the reference sentence (Sign Test with p-value > 0.4). On the accuracy metric, our CP model also performs comparably with the reference as judged by the Sign Test. However, with randomly shuffled content plan, our model (Shuffled CP) fails to match the accuracy of the reference (p-value < 0.05). Our analysis is that the random content plans could contain patterns that are rare or unseen during training. In such cases, our model might fail to produce results that precisely follow the content plan, resulting in a lower accuracy score. Nonetheless, the human results suggest that, while being able to produce fluent and correct sentences, our model is also highly controllable. 
Finally, we note that on the accuracy metric, even the reference sentence does not score a perfect $2.0$. This suggests that our simple heuristic delexicalizer $\mathcal{F}$ introduced in \cref{sec:data_collection} still lapses behind human performance. We leave to future work of designing better $\mathcal{F}$.

\begin{table}[tb]
    \small
	\centering  
	\renewcommand{\arraystretch}{1.2}
	\scalebox{0.785}{
	\begin{tabular}{cccccc}
		\hlinewd{0.75pt}
		\multicolumn{2}{c}{\multirow{2}{*}{\textbf{Model}}}&\multicolumn{2}{c}{\textbf{Quality}}&\multicolumn{2}{c}{\textbf{Diversity}}\\
		\cmidrule(lr){3-4}
		\cmidrule(lr){5-6}
		&&BLEU&PARENT&Self-BLEU$\downarrow$&iBLEU\\
		\hline
		\multirow{4}{*}{B2B}&Greedy&44.15&53.08&100.00&15.32\\
		&Beam&41.58&49.87&75.04&18.26\\
		&Top-$k$&42.47&50.43&82.20&17.54\\
		&Nucleus&42.92&50.91&84.26&17.48\\
		\hline
		\multirow{4}{*}{T5-3B}&Greedy&48.43&57.80&100.00&18.74\\
		&Beam&45.12&55.20&83.68&19.36\\
		&Top-$k$&46.31&55.90&88.86&19.28\\
		&Nucleus&46.53&56.30&90.11&19.20\\
		\hline
		\makecell[c]{\textbf{Ours}}&\multicolumn{4}{l}{}\\
		\hline
		\multirow{2}{*}{Predict}&CP&\textbf{49.10}&\textbf{58.27}&100.00&19.28\\
		&Shuffled CP&40.75&51.96&\textbf{25.91}&\textbf{27.42}\\
		\hline
		\multirow{2}{*}{Oracle}&CP&54.43&62.75&100.00&23.54\\
		&Shuffled CP&42.99&56.17&26.90&29.01\\
		\hlinewd{0.75pt}
	\end{tabular}}
    \caption{Experimental results on the overall ToTTo validation set, where $\downarrow$ means lower is better.}
	\label{tb:diversity_evaluation}
\end{table}

\section{Further Analysis}
In this section, we present and discuss more empirical analyses of the proposed model.

\subsection{Evaluation on Generation Diversity}
\paragraph{Setup.} We first evaluate the ability of different models in generating diverse results on the overall ToTTo validation set. We compare our model with two strong baselines, BERT-to-BERT (B2B) and T5-3B. Given the input data, the baseline models generate the results with different decoding strategies\footnote{For each decoding strategy, five results are generated.
}, including greedy search, beam search (beam size of $10$), top-$k$ sampling ($k=50$) \cite{DBLP:conf/acl/LewisDF18}, and Nucleus sampling ($p=0.9$) \cite{DBLP:conf/iclr/HoltzmanBDFC20}. For our model, to generate diverse results, we simply vary the input content plan and use greedy decoding. We use two variants of the input content plan: (1) the content plan predicted by the content planner (Predict), or (2) the reference content plan (Oracle). For each variant, five results are generated by either using the input content plan (CP), or using five randomly shuffled forms of the content plan (Shuffled CP). The outputs are expected to vary in the latter case only.

\paragraph{Metric.} To measure the output quality, BLEU and PARENT scores are reported. To evaluate the generation diversity, we use Self-BLEU  \cite{DBLP:conf/sigir/ZhuLZGZWY18} and iBLEU \cite{DBLP:conf/acl/SunZ12} metrics\footnote{For all evaluation metrics, we use the same hyper-parameters as in the original works that proposed the metric.}.

\paragraph{Results.} Table \ref{tb:diversity_evaluation} lists the results in which our model ranks best on all metrics. On the quality metrics, we observe notable performance improvements from our model by using the reference content plan (Oracle), suggesting that the choice of content plan has a significant impact on the outputs. By shuffling the content plan, our model shows the largest decrease in BLEU and PARENT, showing that the variation of content plan encourages our model to produce diverse results that have different structures than the reference.

Furthermore, 
we see that, even with different decoding strategies, the baseline models still generate results that are very similar to the ones acquired from greedy search, with their BLEU and PARENT scores relatively unchanged. The results on the diversity metrics also verify the superiority of our model which outperforms the strong T5-3B model by over $57$ and $8$ points on Self-BLEU and iBLEU\footnote{By definition, models using greedy search get 100 Self-BLEU as the generated results are always the same.}. The performance gains suggest that the controllable property of our model is beneficial in producing high-quality as well as diverse results.
\begin{table}[tb]
    \small
	\centering  
	\renewcommand{\arraystretch}{1.2}
	\scalebox{0.88}{
	\begin{tabular}{ccccccc}
		\hlinewd{0.75pt}
        \textbf{Model}&CP&RL&Type&BLEU&PARENT&S-BLEU\\
        \hline
        \textbf{1}&$\times$&$\times$&-&47.50&56.92&43.87\\
        \hline
        \textbf{2}&$\times$&\checkmark&-&48.10&57.34&48.93\\
        \hline
        \multirow{2}{*}{\textbf{3}}&\multirow{2}{*}{\checkmark}&\multirow{2}{*}{$\times$}&Predict&48.53&57.87&57.92\\
        &&&Oracle&53.82&61.99&75.59\\
        \hline
        \multirow{2}{*}{\textbf{Ours}}&\multirow{2}{*}{\checkmark}&\multirow{2}{*}{\checkmark}&Predict&\textbf{49.10}&\textbf{58.27}&\textbf{62.27}\\
        &&&Oracle&54.43&62.75&80.32\\
		\hlinewd{0.75pt}
	\end{tabular}}
    \caption{Ablation Studies on the overall ToTTo validation set. Model 1 gives a baseline for the BART model.}
	\label{tb:ablation_study}
\end{table}

\subsection{Ablation Study}
In this part, we evaluate the importance of each component of our model on the overall ToTTo validation set. Specifically, we study the effect of content plan (CP) and the RL training by removing them iteratively. In addition to BLEU and PARENT, we measure the structure of the model output against the reference content plan with a S-BLEU metric. Given the data $T$, the reference content plan $C$, and the model output $S^{\prime}$, S-BLEU is defined as $B(C, C^{\prime})$, where $B(\cdot, \cdot)$ measures the BLEU score, $C^{\prime}=\mathcal{F}(T,S^{\prime})$, and  $\mathcal{F}$ is the heuristic delexicalizer described in \cref{sec:data_collection}. The results are listed in Table \ref{tb:ablation_study} with the first row showing the baseline results of BART model.

\paragraph{Necessity of Content Plan.} By comparing models with and without the content plan (model 1 vs. 3 and model 2 vs. ours), we observe that the content plan is an effective guiding signal that leads to better results. 
Moreover, we see that the Oracle results outperform the Predict results by a large margin, showing that the quality of the content plan is an important factor of the model performance and future research can focus more on this aspect.

\begin{table*}[t]
    \small
	\centering  
	\renewcommand{\arraystretch}{1.12}
	\scalebox{0.81}{
	\begin{tabular}{ll}
		\hlinewd{0.75pt}
		\multicolumn{2}{c}{\makecell[l]{\textbf{Table}: \textbf{Title}[George Washington Colonials football]  \textbf{Date}[1956] \textbf{Game}[Sun Bowl]
		\textbf{Result}[W 13-0] \textbf{Opponent}[Texas Western] \textbf{Notes}[Bowl Games]}}\\
        \hline
        \multicolumn{2}{c}{\makecell[l]{\textbf{Reference}: In 1956, George Washington Colonials scored 13–0 against Texas Western at the Sun Bowl.}}\\
        
        \hline
        \makecell[c]{\textbf{T5-3B: }\textit{{Greedy Search}}}&\makecell[c]{\textbf{Ours:} \textit{CP}}\\
        \hline
        \multirow{3}{*}{\makecell[l]{George Washington Colonials football won the Sun Bowl (1956) \\over Texas Western.}}&\makecell[l]{\textbf{ICP}: \textcolor{red}{Date} $\rightarrow$ \textcolor{blue}{Title} $\rightarrow$ \textcolor{teal}{Result} $\rightarrow$ \textcolor{orange}{Opponent} $\rightarrow$ \textcolor{violet}{Game}}\\
        &In \textcolor{red}{1956}, \textcolor{blue}{George Washington Colonials football} team scored \textcolor{teal}{13–0} against \textcolor{orange}{Texas}\\
        &\makecell[l]{\textcolor{orange}{Western} in the \textcolor{violet}{Sun Bowl}.}\\
        
        \hline
        \makecell[c]{\textbf{T5-3B: }\textit{{Beam Search}}}&\makecell[c]{\textbf{Ours:} \textit{Shuffled CP}}\\
        \hline
        \multirow{3}{*}{\makecell[l]{\textbf{1:} George Washington Colonials football won the 1956 Sun Bowl\\
        \makecell[l]{\; \;  against Texas Western.}}}&\makecell[l]{\textbf{ICP}: \textcolor{red}{Date} $\rightarrow$  \textcolor{teal}{Result} $\rightarrow$ \textcolor{orange}{Opponent} $\rightarrow$ \textcolor{blue}{Title} $\rightarrow$ \textcolor{violet}{Game}}\\
        &In \textcolor{red}{1956}, with a \textcolor{teal}{13–0} victory over \textcolor{orange}{Texas Western}, the \textcolor{blue}{Colonials football} \textcolor{blue}{team}\\
        &\makecell[l]{ won the \textcolor{violet}{Sun Bowl}.}\\
        
        \hline
        \multirow{3}{*}{\makecell[l]{\textbf{2:} George Washington Colonials won the 1956 Sun Bowl against\\
        \makecell[l]{\; \; Texas Western.}}}&\makecell[l]{\textbf{ICP}: \textcolor{blue}{Title} $\rightarrow$ \textcolor{violet}{Game} $\rightarrow$ \textcolor{red}{Date} $\rightarrow$ \textcolor{teal}{Result} $\rightarrow$ \textcolor{orange}{Opponent}}\\
        &\textcolor{blue}{George Washington Colonials football} won the \textcolor{violet}{Sun} \textcolor{violet}{Bowl} in \textcolor{red}{1956} with a \textcolor{teal}{13–0}\\
        &\makecell[l]{  victory over \textcolor{orange}{Texas Western}.}\\
        
        \hline
       \multirow{3}{*}{\makecell[l]{\textbf{3:} In 1956, George Washington Colonials won the Sun Bowl against\\
       \makecell[l]{\; \;  Texas Western.}}}&\makecell[l]{\textbf{ICP}: \textcolor{violet}{Game} $\rightarrow$ \textcolor{teal}{Result} $\rightarrow$ \textcolor{blue}{Title} $\rightarrow$ \textcolor{orange}{Opponent} $\rightarrow$ \textcolor{red}{Date}}\\
        &In the \textcolor{violet}{Sun Bowl}, a \textcolor{teal}{13–0} victory for \textcolor{blue}{George Washington} \textcolor{blue}{Colonials} over \textcolor{orange}{Texas}\\
        &\makecell[l]{ \textcolor{orange}{Western} in \textcolor{red}{1956}.}\\

        \hline
        \multirow{3}{*}{\makecell[l]{\textbf{4:} George Washington Colonials won the  Sun Bowl against Texas\\
        \makecell[l]{\; \;  Western in 1956.}}}&\makecell[l]{\textbf{ICP}: \textcolor{blue}{Title} $\rightarrow$ \textcolor{orange}{Opponent} $\rightarrow$ \textcolor{violet}{Game} $\rightarrow$ \textcolor{teal}{Result} $\rightarrow$ \textcolor{red}{Date}}\\
        &\textcolor{blue}{George Washington Colonials football} team defeated \textcolor{orange}{Texas} \textcolor{orange}{Western} in the \textcolor{violet}{Sun}\\
        &\makecell[l]{\textcolor{violet}{Bowl}, with \textcolor{teal}{13–0}, in \textcolor{red}{1956}.}\\
        \hline
        \multirow{3}{*}{\makecell[l]{\textbf{5:} George Washington Colonials football won the 1956 Sun Bowl\\
        \makecell[l]{\; \;  over Texas Western.}}}&\makecell[l]{\textbf{ICP}: \textcolor{orange}{Opponent} $\rightarrow$  \textcolor{violet}{Game} $\rightarrow$ \textcolor{red}{Date} $\rightarrow$ \textcolor{teal}{Result} $\rightarrow$ \textcolor{blue}{Title}}\\
        &The \textcolor{blue}{Colonials} defeated \textcolor{orange}{Texas Western} in the \textcolor{violet}{Sun Bowl} \textcolor{red}{1956}, with a \textcolor{teal}{13–0} score,\\
        &\makecell[l]{  by \textcolor{blue}{George Washington Colonials}. \; }\\
		\hlinewd{0.75pt}
	\end{tabular}}
    \caption{Case study on ToTTo dataset. Given the input data, we present the generated results from various models using different decoding strategies. \textbf{ICP} denotes the ``input content plan". (Best viewed in color)}
	\label{tb:case_study}
\end{table*}

\begin{table*}[t]
    \small
	\centering  
	\renewcommand{\arraystretch}{1.2}
	\scalebox{0.78}{
	\begin{tabular}{cl}
	\hlinewd{0.75pt}
	\textbf{Tripleset}&\makecell[l]{(Alan Bean | nationality | United States),	(Alan Bean | occupation | Test pilot), (Alan Bean | birthPlace | Wheeler , Texas),\\ (Alan Bean | selectedByNASA | 1963), (Alan Bean | status | "Retired")}\\
	\hlinewd{0.75pt}
	\textbf{Reference} & Alan Bean is a US national born in Wheeler, Texas. He is a retired test pilot who joined NASA in 1963.\\
	\hline
	\multirow{6}{*}{\makecell[c]{\textbf{Ours}\\(Shuffled CP)}}&\makecell[l]{\textbf{ICP}: \textcolor{red}{nationality} $\rightarrow$ \textcolor{blue}{birthPlace} $\rightarrow$ \textcolor{teal}{selectedByNASA} $\rightarrow$	\textcolor{orange}{status} $\rightarrow$ \textcolor{violet}{occupation}}\\
	&\makecell[l]{Alan Bean is a \textcolor{red}{US national} who was born in \textcolor{blue}{Wheeler, Texas}. He was selected by NASA in \textcolor{teal}{1963} and is now \textcolor{orange}{retired}. He was a \textcolor{violet}{test pilot}.}\\
	\cline{2-2}
	&\makecell[l]{\textbf{ICP}: \textcolor{red}{nationality} $\rightarrow$ \textcolor{violet}{occupation} $\rightarrow$ \textcolor{teal}{selectedByNASA} $\rightarrow$ \textcolor{blue}{birthPlace} $\rightarrow$ \textcolor{orange}{status}}\\
	&Alan Bean is a \textcolor{red}{US national} who served as a \textcolor{violet}{test pilot} and was selected by NASA in \textcolor{teal}{1963}. He was born in Wheeler, Texas and is now \textcolor{orange}{retired}.\\
	\cline{2-2}
	&\makecell[l]{\textbf{ICP}: \textcolor{teal}{selectedByNASA} $\rightarrow$ \textcolor{violet}{occupation} $\rightarrow$ \textcolor{orange}{status} $\rightarrow$ \textcolor{blue}{birthPlace} $\rightarrow$ \textcolor{red}{nationality}}\\
	&Alan Bean was selected by NASA in \textcolor{teal}{1963} as a \textcolor{violet}{test pilot}. He is now \textcolor{orange}{retired}. He was born in \textcolor{blue}{Wheeler, Texas} and is a \textcolor{red}{United States national}.\\
	\hlinewd{0.75pt}
	\end{tabular}}
    \caption{Case study of our model's results on WebNLG dataset. (best viewed in color)}
	\label{tb:webnlg_case_study}
\end{table*}

\paragraph{Effect of RL.} By comparing the models trained with and without RL  (model 1 vs. 2 and model 3 vs. ours), we see that training with our proposed RL objective consistently improves the model performance. The most notable improvement is observed in S-BLEU which means that the generated outputs better follow the input content plan. This is in line with our hypothesis that our reward function in Eq. \eqref{eq:reward} helps to improve the model's adherence to the output structure defined by the content plan. 

\subsection{Case Study}
\label{sec:case_study}
To gain more insights into our model, we present generated examples from ToTTo and WebNLG datasets\footnote{More examples are shown in the Appendix \ref{sec:more_example}.} 
in Table \ref{tb:case_study} and Table \ref{tb:webnlg_case_study}, respectively.

\paragraph{Quality.} In Table \ref{tb:case_study}, we compare our model with predicted content plan against T5-3B. 
We see that T5-3B fails to produce the key game result (i.e. \textit{13-0}) in its outputs. In contrast, by following the content plan, our model is able to maintain all key information in its generated results.


\paragraph{Diversity and Controllability.} Next, we examine the output diversity and controllability. For the T5-3B model, when using beam search, only the position of the term \textit{``1956''} varies, showing its reduced ability to generate diverse outputs. For our model, the variation of content plan leads to outputs with diverse structures. Furthermore, the results show that our model is not only able to control the intra-sentence output structure as shown in Table \ref{tb:case_study} but also to control the inter-sentence output structure as shown in Table \ref{tb:webnlg_case_study}. 

\paragraph{Error Analysis.} 
We show one failure case in the bottom right cell of Table \ref{tb:case_study}, in which it repeats the \textit{Title} key twice in the output.
Our analysis for such error is that the randomly shuffled content plan might contain patterns that are rarely seen in training. 
One possible solution is filtering out rare content plan patterns via statistical approaches such as bigram statistics.

\section{Conclusion}

In this study, we propose a new Plan-then-Generate (PlanGen) framework for data-to-text generation which can be easily applied to data with different structures. Extensive experiments and analyses are conducted on two benchmark datasets. Both automatic and human evaluation results demonstrate that our model is highly controllable. Furthermore, compared with previous studies, our model achieves better results both in terms of the generation quality as well as the output diversity. Our code, models and other related resources can be found in  \url{https://github.com/yxuansu/PlanGen/}

\section*{Acknowledgments}
The authors wish to thank Ehsan Shareghi, Zaiqiao Meng, Piji Li, and Benjamin Muller for their insightful discussions and support. Many thanks to our anonymous reviewers for their suggestions and comments.


\bibliography{anthology,custom}
\bibliographystyle{acl_natbib}

\clearpage
\onecolumn
\appendix
\section{Details of Human Evaluation Setup}
\label{sec:human_evaluation_detail}
\begin{figure*}[h] 
	\centering    
	\setlength{\abovecaptionskip}{3pt}
\includegraphics[width=0.99\textwidth]{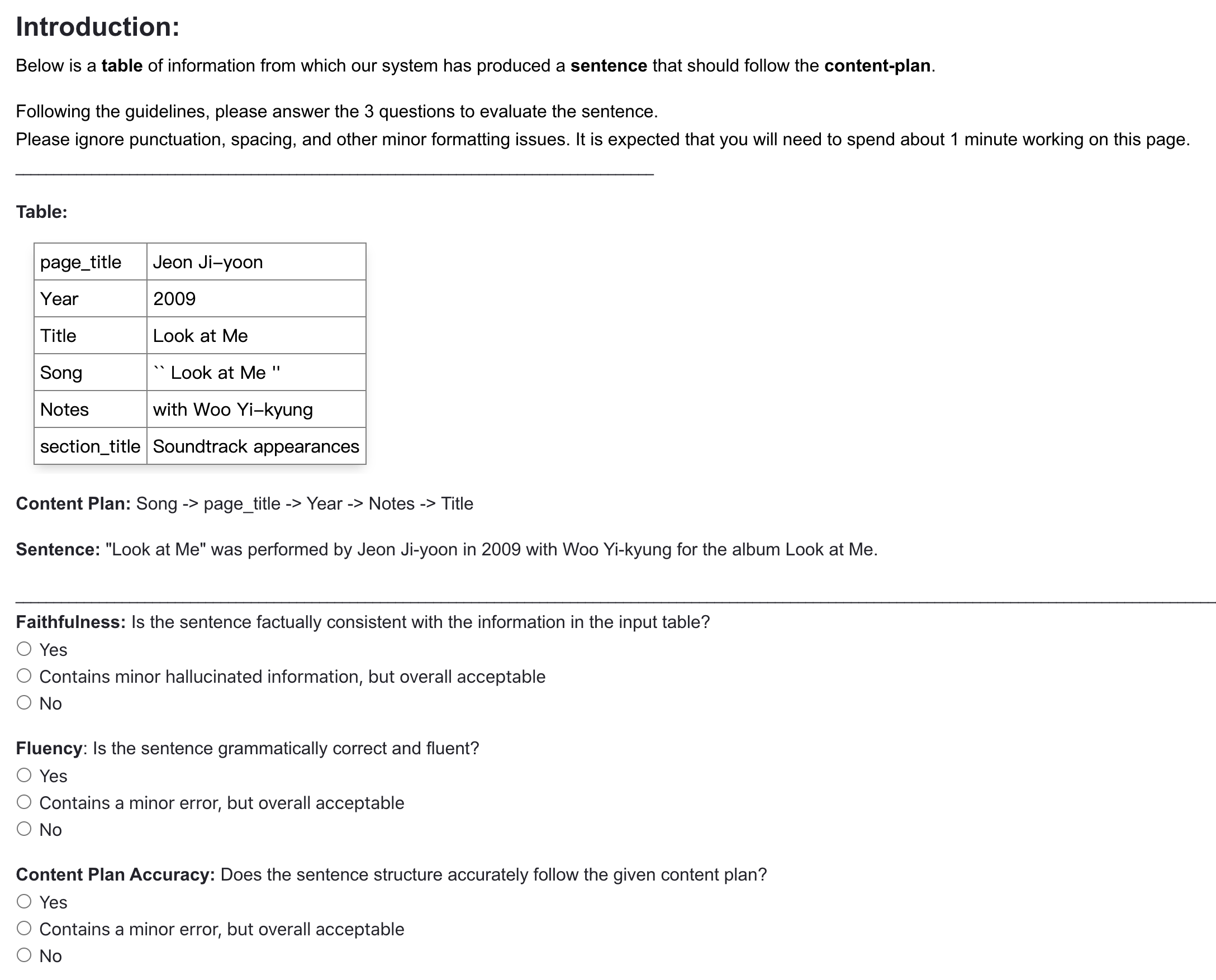}
    \caption{Example of Human Evaluation Interface}
    \label{fig:UI}
\end{figure*}

To perform human evaluation, we randomly sample 200 samples from the ToTTo validation set. For each sampled data, we use each baseline model (BERT-to-BERT and T5-3B) to produce one result. As for our model, we produce 6 different results (one with the predicted content plan, the other five with five randomly shuffled versions of the predicted content plan). Therefore, for each case, we have 9 different results (1 from BERT-to-BERT, 1 from T5-3B, 6 from our model, and 1 reference). To reduce human bias, we randomly shuffle these 1800 data points before presenting them to three annotators. Each annotator is asked to assess all these 1800 data points. Because BERT-to-BERT and T5-3B do not take the content plan as input, thus we only measure the accuracy score for the results generated by our model and the reference sentence. Note that the accuracy score of the reference sentence is measured against the reference content plan. In Figure  \ref{fig:UI}, we show an example of the human evaluation interface. 

\clearpage
\section{More Examples of Generated Result}
\label{sec:more_example}
In this part, we provide more generated examples of our model. The generated results on samples from WebNLG and ToTTo datasets are shown in Table \ref{tb:webnlg_case_study_2} and \ref{tb:totto_example}, respectively.
From the results, we can see that our model is able to generate fluent and diverse sentence while maintaining the structure defined by the desired content plan. In particular, our model is able to control the output structure both on the inter-sentence level (i.e. the structure across multiple sentences) as shown in Table \ref{tb:webnlg_case_study} and on the intra-sentence level (i.e. the structure within a single sentence) as shown in Table \ref{tb:totto_example}. These results further demonstrate the applicability and generalization ability of our model.

\begin{table*}[h]
    \small
	\centering  
	\renewcommand{\arraystretch}{1.4}
	\scalebox{0.95}{
	\begin{tabular}{cl}
	\hlinewd{0.75pt}
	\textbf{Tripleset}&\makecell[l]{(Allama Iqbal International Airport | location | Pakistan),\\ (Allama Iqbal International Airport | runwayLength | 2900.0),\\ (Allama Iqbal International Airport | cityServed | Lahore),\\ 
	(Allama Iqbal International Airport | operatingOrganisation | Pakistan Civil Aviation Authority)}\\
	\hline
	\textbf{Reference}&\makecell[l]{Allama Iqbal International Airport is located in Lahore at Pakistan. It has a runway length of 2900\\ and is operated by the Pakistan Civil Aviation Authority.}\\
	\hline
	\multirow{10}{*}{\makecell[c]{\textbf{Ours}\\(Shuffled CP)}}&\makecell[l]{\textbf{ICP}: \textcolor{red}{cityServed} $\rightarrow$	\textcolor{blue}{location} $\rightarrow$	\textcolor{violet}{runwayLength} $\rightarrow$ \textcolor{teal}{operatingOrganisation}}\\
	&\makecell[l]{Allama Iqbal International Airport serves the city of \textcolor{red}{Lahore} and is located in \textcolor{blue}{Pakistan}. The runway\\ length is \textcolor{violet}{2900.0} and the airport is operated by the \textcolor{teal}{Pakistan Civil Aviation Authority}.}\\
	\cline{2-2}
	&\makecell[l]{\textbf{ICP}: \textcolor{teal}{operatingOrganisation}	$\rightarrow$ \textcolor{red}{cityServed}	$\rightarrow$ \textcolor{blue}{location}	$\rightarrow$ \textcolor{violet}{runwayLength}}\\
	&\makecell[l]{The \textcolor{teal}{Pakistan Civil Aviation Authority} is the operating organisation of the Allama Iqbal International\\ Airport which serves the city of \textcolor{red}{Lahore} in \textcolor{blue}{Pakistan}. The airport has a runway length of \textcolor{violet}{2900.0}.}\\
	\cline{2-2}
	&\makecell[l]{\textbf{ICP}: \textcolor{violet}{runwayLength} $\rightarrow$ \textcolor{red}{cityServed} $\rightarrow$ \textcolor{blue}{location} $\rightarrow$ \textcolor{teal}{operatingOrganisation}}\\
	&\makecell[l]{Allama Iqbal International Airport has a runway length of \textcolor{violet}{2900.0} and serves the city of \textcolor{red}{Lahore} in\\ \textcolor{blue}{Pakistan}. The airport is operated by the \textcolor{teal}{Pakistan Civil Aviation Authority}.}\\
	\cline{2-2}
	&\makecell[l]{\textbf{ICP}: \textcolor{blue}{location} $\rightarrow$ \textcolor{teal}{operatingOrganisation} $\rightarrow$ \textcolor{red}{cityServed} $\rightarrow$ \textcolor{violet}{runwayLength}}\\
	&\makecell[l]{Allama Iqbal International Airport is located in \textcolor{blue}{Pakistan} and is operated by the \textcolor{teal}{Pakistan Civil Aviation}\\
	\textcolor{teal}{Authority}. The airport serves the city of \textcolor{red}{Lahore} and has a runway length of \textcolor{violet}{2900.0}.}\\
	\cline{2-2}
	&\makecell[l]{\textbf{ICP}: \textcolor{violet}{runwayLength} $\rightarrow$ \textcolor{teal}{operatingOrganisation} $\rightarrow$ \textcolor{red}{cityServed} $\rightarrow$ \textcolor{blue}{location}}\\
	&\makecell[l]{Allama Iqbal International Airport has a runway length of \textcolor{violet}{2900.0} and is operated by the \textcolor{teal}{Pakistan Civil}\\\textcolor{teal}{Aviation Authority}. The airport serves the city of \textcolor{red}{Lahore} and is located in \textcolor{blue}{Pakistan}.}\\
		\hline
		\hline
	\textbf{Tripleset}&\makecell[l]{(Turkey | leaderTitle | president of Turkey), (Turkey | leader | Ahmet Davutoglu),
	(Turkey | capital | Ankara),\\ (Turkey | largestCity | Istanbul),
	(Atatürk Monument (Izmir) | location | Turkey)}\\
	\hline
	\textbf{Reference}&\makecell[l]{The Ataturk Monument (Izmir) is located in Turkey, whose largest city is Istanbul and whose capital is\\ Ankara. President of Turkey is the official title of the Turkish leader and his name is Ahmet Davutoglu.} \\
	\hline
    \multirow{10}{*}{\makecell[c]{\textbf{Ours}\\(Shuffled CP)}}&\makecell[l]{\textbf{ICP}: \textcolor{red}{location} $\rightarrow$ \textcolor{blue}{leaderTitle} $\rightarrow$ \textcolor{orange}{capital} $\rightarrow$ \textcolor{violet}{leader} $\rightarrow$ \textcolor{cyan}{largestCity}}\\
    &\makecell[l]{The Atatürk Monument (Izmir) is located in \textcolor{red}{Turkey}, where the leader is known as the \textcolor{blue}{president} and the\\ capital city is \textcolor{orange}{Ankara}. \textcolor{violet}{Ahmet Davutoglu} is the leader of Turkey and the largest city is \textcolor{cyan}{Istanbul}.}\\
    \cline{2-2}
    &\makecell[l]{\textbf{ICP}: \textcolor{orange}{capital} $\rightarrow$ \textcolor{blue}{leaderTitle} $\rightarrow$ \textcolor{violet}{leader} $\rightarrow$ \textcolor{cyan}{largestCity} $\rightarrow$ \textcolor{red}{location}}\\
    &\makecell[l]{\textcolor{orange}{Ankara} is the capital of Turkey, which is lead by \textcolor{blue}{president}  \textcolor{violet}{Ahmet Davutoglu}. The largest city is \textcolor{cyan}{Istanbul}\\ and the Atatürk Monument is located in \textcolor{red}{Turkey}.}\\
    \cline{2-2}
    &\makecell[l]{\textbf{ICP}: \textcolor{cyan}{largestCity} $\rightarrow$ \textcolor{red}{location} $\rightarrow$ \textcolor{orange}{capital} $\rightarrow$ \textcolor{blue}{leaderTitle} $\rightarrow$ \textcolor{violet}{leader}}\\
    &\makecell[l]{\textcolor{cyan}{Istanbul} is the largest city in \textcolor{red}{Turkey}, where the Ataturk Monument is located and the capital is \textcolor{orange}{Ankara}. \\The leader of Turkey is known as the \textcolor{blue}{president} and \textcolor{violet}{Ahmet Davutoglu} is the leader.}\\
    \cline{2-2}
    &\makecell[l]{\textbf{ICP}: \textcolor{red}{location} $\rightarrow$ \textcolor{orange}{capital} $\rightarrow$ \textcolor{violet}{leader} $\rightarrow$ \textcolor{cyan}{largestCity} $\rightarrow$ \textcolor{blue}{leaderTitle}}\\
    &\makecell[l]{The Atatürk Monument (Izmir) is located in \textcolor{red}{Turkey}, where the capital is \textcolor{orange}{Ankara}, the leader is \textcolor{violet}{Ahmet}\\ \textcolor{violet}{Davutoglu} and the largest city is \textcolor{cyan}{Istanbul}. The leader of Turkey is known as the \textcolor{blue}{president of Turkey}.}\\
    \cline{2-2}
    &\makecell[l]{\textbf{ICP}: \textcolor{red}{location} $\rightarrow$ \textcolor{cyan}{largestCity} $\rightarrow$ \textcolor{orange}{capital} $\rightarrow$ \textcolor{blue}{leaderTitle} $\rightarrow$ \textcolor{violet}{leader}}\\
    &\makecell[l]{The Atatürk Monument (Izmir) is located in \textcolor{red}{Turkey}, where the largest city is \textcolor{cyan}{Istanbul} and the capital\\ is \textcolor{orange}{Ankara}. The leader of Turkey is known as the \textcolor{blue}{president} and \textcolor{violet}{Ahmet Davutoglu} is the leader.}\\
	\hlinewd{0.75pt}
	\end{tabular}}
    \caption{Examples of generated result from WebNLG dataset, where \textbf{ICP} denotes ``input content plan". The expressions correspond to different contents are highlighted with different colors. (best viewed in color)}
	\label{tb:webnlg_case_study_2}
\end{table*}

\begin{table*}[t]
    \small
	\centering  
	\renewcommand{\arraystretch}{1.12}
	\scalebox{0.93}{
	\begin{tabular}{l}
		\hlinewd{0.75pt}
		\makecell[c]{\textbf{Input Table}}\\
		\hline
		\makecell[c]{\textbf{Page_Title}[List of New Zealand records in swimming] \textbf{Event}[400m freestyle] \textbf{Time}[4:03.63] \textbf{Name}[Lauren Boyle]\\
		\textbf{Club}[New Zealand] \textbf{Date}[29 July 2012] \textbf{Meet}[Olympic Games] \textbf{Location}[London, United Kingdom] \textbf{Section_Title}[Women]}\\
		\hline
		\multicolumn{1}{c}{\textit{Reference Sentence}}\\
		\hline
		At the 2012 Olympics in London, Boyle was fourth fastest in the heats of the 400m freestyle in a New Zealand record 4:03.63.\\
		\hline
		\multicolumn{1}{c}{\textit{Controlled Data-to-Text Generation}}\\
		\hline
		\textbf{ICP}: \textcolor{red}{Date} $\rightarrow$ \textcolor{blue}{Name} $\rightarrow$ \textcolor{orange}{Event} $\rightarrow$ \textcolor{teal}{Time} $\rightarrow$ \textcolor{violet}{Location} $\rightarrow$ \textcolor{cyan}{Meet} $\rightarrow$ \textcolor{brown}{Page_Title}\\
		On \textcolor{red}{29 July 2012}, New Zealand's \textcolor{blue}{Lauren Boyle} finished the \textcolor{orange}{400 metre freestyle} in \textcolor{teal}{4:03.63} at \textcolor{violet}{London} \textcolor{cyan}{Olympics}, which was a \\\textcolor{brown}{New Zealand record}.\\
		\hline
		\textbf{ICP}: \textcolor{brown}{Page_Title} $\rightarrow$ \textcolor{violet}{Location} $\rightarrow$ \textcolor{cyan}{Meet} $\rightarrow$ \textcolor{blue}{Name} $\rightarrow$ \textcolor{orange}{Event} $\rightarrow$ \textcolor{teal}{Time} $\rightarrow$ \textcolor{red}{Date}\\
        The \textcolor{brown}{New Zealand swimming record} was set in \textcolor{violet}{London} at the 2012 London \textcolor{cyan}{Olympics}, where \textcolor{blue}{Lauren Boyle} finished the \textcolor{orange}{400 metre}\\ \textcolor{orange}{freestyle} in \textcolor{teal}{4:03.63}, in \textcolor{red}{July 2012}\\
        \hline
        \textbf{ICP}: \textcolor{cyan}{Meet} $\rightarrow$ \textcolor{violet}{Location} $\rightarrow$ \textcolor{red}{Date} $\rightarrow$ \textcolor{blue}{Name} $\rightarrow$ \textcolor{teal}{Time} $\rightarrow$ \textcolor{orange}{Event} $\rightarrow$ Page_Title\\
        At the \textcolor{cyan}{Olympic Games} in \textcolor{violet}{London} in \textcolor{red}{July 2012}, \textcolor{blue}{Lauren Boyle} finished in \textcolor{teal}{4:03.63} in the \textcolor{orange}{400 metre freestyle}, setting a \textcolor{brown}{New Zealand}\\  \textcolor{brown}{record}.\\
        \hline
        \textbf{ICP}: \textcolor{blue}{Name} $\rightarrow$ \textcolor{orange}{Event} $\rightarrow$ \textcolor{red}{Date} $\rightarrow$ \textcolor{cyan}{Meet} $\rightarrow$ \textcolor{violet}{Location} $\rightarrow$ \textcolor{brown}{Page_Title} $\rightarrow$ \textcolor{teal}{Time}\\
        \textcolor{blue}{Lauren Boyle} competed in the \textcolor{orange}{400 metre freestyle} at the \textcolor{red}{2012} \textcolor{cyan}{Olympic Games} in \textcolor{violet}{London}, finishing with a \textcolor{brown}{New Zealand record}\\ time of \textcolor{teal}{4:03.63}.\\
        \hline
        \textbf{ICP}: \textcolor{violet}{Location} $\rightarrow$ \textcolor{orange}{Event} $\rightarrow$ \textcolor{blue}{Name} $\rightarrow$ \textcolor{teal}{Time} $\rightarrow$ \textcolor{brown}{Page_Title} $\rightarrow$ \textcolor{red}{Date} $\rightarrow$ \textcolor{cyan}{Meet}\\
        In \textcolor{violet}{London}, in the \textcolor{orange}{400m freestyle}, \textcolor{blue}{Lauren Boyle} finished in \textcolor{teal}{4:03.63}, setting a \textcolor{brown}{New Zealand record} at the \textcolor{red}{2012} \textcolor{cyan}{Olympic Games}.\\
        \hline
        \hline
		\makecell[c]{\textbf{Input Table}}\\
		\hline
		\makecell[c]{\textbf{Page_Title}[Kazunari Murakami] \textbf{Result}[Loss] \textbf{Opponent}[Akihiro Gono] \textbf{Method}[KO (head kick)] \\\textbf{Date}[October 13, 1995]
		\textbf{Event}[Lumax Cup: Tournament of J '95] \textbf{Section_Title}[Mixed martial arts record]}\\
		\hline
		\multicolumn{1}{c}{\textit{Reference Sentence}}\\
		\hline
		Murakami began his career fighting in the Lumax Cup: Tournament of J '95 on October 13, 1995, where he lost his first fight to\\ Akihiro Gono by knockout.\\
		\hline
		\multicolumn{1}{c}{\textit{Controlled Data-to-Text Generation}}\\
		\hline
		\textbf{ICP}: \textcolor{red}{Page_Title} $\rightarrow$ \textcolor{blue}{Date} $\rightarrow$ \textcolor{violet}{Event} $\rightarrow$ \textcolor{teal}{Opponent}\\
		\textcolor{red}{Kazunari Murakami} made his debut on \textcolor{blue}{October 13, 1995} at \textcolor{violet}{Lumax Cup: Tournament of J '95}, losing to \textcolor{teal}{Akihiro Gono} by KO.\\
		\hline
		\textbf{ICP}: \textcolor{violet}{Event} $\rightarrow$ \textcolor{blue}{Date} $\rightarrow$ \textcolor{red}{Page_Title} $\rightarrow$ \textcolor{teal}{Opponent}\\
		\textcolor{violet}{At Lumax Cup: Tournament of J '95} on \textcolor{blue}{October 13, 1995}, \textcolor{red}{Kazunari Murakami} lost to \textcolor{teal}{Akihiro Gono} by KO.\\
		\hline
		\textbf{ICP}: \textcolor{teal}{Opponent} $\rightarrow$ \textcolor{red}{Page_Tilte} $\rightarrow$ \textcolor{violet}{Event} $\rightarrow$ \textcolor{blue}{Date}\\
		\textcolor{teal}{Akihiro Gono} defeated \textcolor{red}{Kazunari Murakami} at \textcolor{violet}{Lumax Cup: Tournament of J '95} on \textcolor{blue}{October 13, 1995}.\\
		\hline
		\textbf{ICP}: \textcolor{blue}{Date} $\rightarrow$ \textcolor{teal}{Opponent} $\rightarrow$ \textcolor{red}{Page_Title} $\rightarrow$ \textcolor{violet}{Event}\\
		\textcolor{blue}{On October 13, 1995}, \textcolor{teal}{Akihiro Gono} defeated \textcolor{red}{Kazunari Murakami} at \textcolor{violet}{Lumax Cup: Tournament of J '95}.\\
		\hline
		\textbf{ICP}: \textcolor{violet}{Event} $\rightarrow$ \textcolor{red}{Page_Title} $\rightarrow$ \textcolor{teal}{Opponent} $\rightarrow$ \textcolor{blue}{Date}\\
		At \textcolor{violet}{Lumax Cup: Tournament of J '95}, \textcolor{red}{Kazunari Murakami} lost to \textcolor{teal}{Akihiro Gono} by KO on \textcolor{blue}{October 13, 1995}.\\
		\hline
		\hline
		\makecell[c]{\textbf{Input Table}}\\
		\hline
		\makecell[c]{\textbf{Page_Title}[Reform Party of the United States of America] \textbf{Year}[2008] \\\textbf{Name}[Frank McEnulty] \textbf{Home_state}[California] 
		\textbf{Section_Title}[Presidential tickets]}\\
		\hline
		\multicolumn{1}{c}{\textit{Reference Sentence}}\\
		\hline
		Frank McEnulty of California, was nominated to be the Reform Party's 2008 presidential candidate.\\
		\hline
		\multicolumn{1}{c}{\textit{Controlled Data-to-Text Generation}}\\
		\hline
		\textbf{ICP}: \textcolor{red}{Year} $\rightarrow$ \textcolor{blue}{Page_Title} $\rightarrow$ \textcolor{violet}{Name} $\rightarrow$ \textcolor{teal}{Home_state} $\rightarrow$ \textcolor{orange}{Section_Title}\\
		In \textcolor{red}{2008}, the \textcolor{blue}{Reform Party of the United States of America} nominated \textcolor{violet}{Frank McEnulty} of \textcolor{teal}{California} as its \textcolor{orange}{presidential candidate}.\\
		\hline
		\textbf{ICP}: \textcolor{blue}{Page_Title} $\rightarrow$ \textcolor{orange}{Section_Title} $\rightarrow$ \textcolor{teal}{Home_state} $\rightarrow$ \textcolor{red}{Year} $\rightarrow$ \textcolor{violet}{Name}\\
		\textcolor{blue}{Reform Party of the United States of America} nominated its first \textcolor{orange}{presidential nominee} from \textcolor{teal}{California} in \textcolor{red}{2008}, \textcolor{violet}{Frank McEnulty}.\\
		\hline
		\textbf{ICP}: \textcolor{teal}{Home_state} $\rightarrow$ \textcolor{violet}{Name} $\rightarrow$ \textcolor{orange}{Section_Title} $\rightarrow$ \textcolor{red}{Year} $\rightarrow$ \textcolor{blue}{Page_Title}\\
		\textcolor{teal}{California}'s \textcolor{violet}{Frank McEnulty} was nominated as \textcolor{orange}{presidential candidate} in \textcolor{red}{2008} by the \textcolor{blue}{Reform Party of the United States of America}.\\
		\hline
		\textbf{ICP}: \textcolor{blue}{Page_Title} $\rightarrow$ \textcolor{violet}{Name} $\rightarrow$ \textcolor{teal}{Home_state} $\rightarrow$ \textcolor{orange}{Section_Title} $\rightarrow$ \textcolor{red}{Year}\\
		\textcolor{blue}{Reform Party of the United States of America} nominated \textcolor{violet}{Frank McEnulty} of \textcolor{teal}{California} as its \textcolor{orange}{presidential candidate} in \textcolor{red}{2008}.\\
		\hline
		\textbf{ICP}: \textcolor{red}{Year} $\rightarrow$ \textcolor{violet}{Name} $\rightarrow$ \textcolor{blue}{Page_Title} $\rightarrow$ \textcolor{teal}{Home_state} $\rightarrow$ \textcolor{orange}{Section_Title}\\
		In \textcolor{red}{2008}, \textcolor{violet}{Frank McEnulty} of \textcolor{blue}{Reform Party of the United States of America} from \textcolor{teal}{California} ran for the \textcolor{orange}{presidential election}.\\
		\hlinewd{0.75pt}
	\end{tabular}}
    \caption{Examples of generated result from ToTTo dataset, where \textbf{ICP} denotes ``input content plan". The expressions correspond to different contents are highlighted with different colors. (best viewed in color)}
	\label{tb:totto_example}
\end{table*}

\end{document}